\documentclass[10pt,twocolumn,letterpaper]{article}

\usepackage[pagenumbers]{cvpr} 

\usepackage{graphicx}
\usepackage{amsmath}
\usepackage{amssymb}
\usepackage{booktabs}
\usepackage{comment}
\usepackage{amsmath}
\usepackage{adjustbox}

\usepackage[pagebackref,breaklinks,colorlinks]{hyperref}

\usepackage[capitalize]{cleveref}
\crefname{section}{Sec.}{Secs.}
\Crefname{section}{Section}{Sections}
\Crefname{table}{Table}{Tables}
\crefname{table}{Tab.}{Tabs.}

\def\cvprPaperID{*****} 
\def\confName{CVPR}
\def\confYear{2024}

\begin{document}

\title{GAD-Generative Learning for HD Map-Free Autonomous Driving}
\author{Weijian Sun, Yanbo Jia, Qi Zeng, Zihao Liu, Jiang Liao, Yue Li, Xianfeng Li\\
JIDU AUTO\\
}
\maketitle


\begin{abstract}

    Deep-learning-based techniques have been widely adopted for autonomous driving software stacks for mass production in recent years, focusing primarily on perception modules, with some work extending this method to prediction modules. However, the downstream planning and control modules are still designed with hefty hand-crafted rules, dominated by optimization-based methods such as quadratic programming or model predictive control. This results in a performance bottleneck for autonomous driving systems in that corner cases simply cannot be solved by enumerating hand-crafted rules. We present a deep-learning-based approach that brings prediction, decision, and planning modules together with the attempt to overcome the rule-based methods' deficiency in real-world applications of autonomous driving, especially for urban scenes.
    The DNN model we proposed is solely trained with 10 hours of valid human driver data and supports all mass-production ADAS features available on the market to date. 
    This method is deployed onto a Jiyue test car without modification to its factory-ready sensor set and compute platform. The feasibility, usability, and commercial potential are demonstrated in this article. 
      
\end{abstract}

\section{Introduction}
\label{sec:intro}

Autonomous driving software on the market to date is usually designed following a classical engineering methodology called ``divide and conquer", where the problem is divided into smaller sub-tasks to reduce complexity. These sub-tasks usually include perception, prediction, motion planning, and control. However, this easily interpretable rule-based method is highly dependent on accurate prior knowledge, which severely restricts its scalability. High-definition (HD) maps, among various forms of prior knowledge, play a crucial part in offering rich semantic information. 
Unfortunately, HD maps are proven hard to scale up due to the complexity and costliness of map generation and maintenance, as well as failure to reflect changes in real-world road conditions timely, which is often constrained by resource-intensive collection processes. 
Furthermore, the rule-based planning module is held accountable for most failures and deficiencies of the autonomous driving system in real driving scenarios, especially for urban scenes.
Hence, it is often regarded as the bottleneck of the entire software stack.  New rules must be designed and re-tuned in accordance with different driving scenarios and new corner cases. This process is expensive and it scales poorly for the simple fact that the scenario space for the Autonomous Driving Vehicle(ADV) is infinite while rules can only be of a limited number. 


\begin{figure}[t]
    \centering
    \includegraphics[width=1\linewidth]{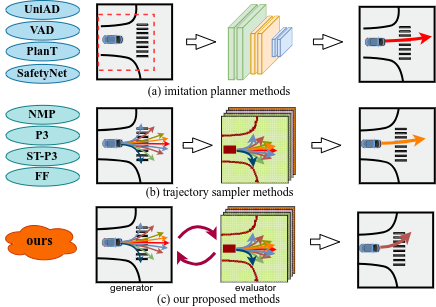}
    \caption{Comparison on the design of model-based planning. (a)Imitation planner with single trajectory output (\cite{hu2023_uniad, jiang2023vad, renz2022plant, vitelli2022safetynet}). (b)Sample a bunch of candidates and evaluate with cost volume(\cite{zeng2021endtoend, casas2021mp3, sadat2020perceive, hu2022st}). (c)Proposed generative method, in which the data-driven generator gets feedback from the evaluator.}
    \label{fig:fig0}
\end{figure}

End-to-end autonomous driving, which aims to consider sub-tasks in a full-stack manner jointly, has been advocated to address the aforementioned challenges. Methods\cite{hu2023_uniad, jiang2023vad, kendall2018learning, shao2022interfuser,mile2022, Pomerleau-1989-15721, codevilla2018endtoend, müller2018driving} that facilitate direct learning from sensor data to trajectories or driving commands obviate the necessity for manually devised heuristic rules and the dependency on HD maps. 
However, these methodologies are deficient in interpretability, which is crucial for elucidating and monitoring the system's evolution, consequently complicating optimization efforts by developers. Moreover, the absence of modularized structure and prior knowledge makes these methods exceedingly brittle when operationalized in real-world closed-loop.\par


To overcome the aforementioned problems, we explore an alternative method that runs entirely on the edge and purely relies on on-the-fly computation of local topology reconstructed with landmarks and occupancy grids \cite{elfes2013occupancy} based on perception results. The reconstruction is designed to support motion planning in urban scenarios, and we explore its practicality for daily commuting first, primarily due to considerations of scalability and mass production feasibility.
However, the proposed methodology is not limited to commuting scenarios; it supports a versatile selection of route formats. We also validated its non-commuting variant in the experiments.

To break the bottleneck brought by rule-based planning, we propose an interpretable and tunable framework to handle the prediction-planning tasks in a data-driven and integrated manner. Our framework consists of a trajectory generator and evaluator. The generator utilizes vectorized embedding from parametric input to perform trajectory generation, while the evaluator leverages rasterized non-parametric information in grid form to validate and score candidates
\footnote{We note that by $\textit{parametric}$ form we mean the structured numerical representations of certain perception outputs, such as the observed vehicle states. Conversely, the $\textit{non-parametric}$ form refers to unstructured representations that cannot be described by parameters, such as occupancy.}. We validate our framework in a closed loop by deploying and testing it in real-world user-level autonomous driving vehicles.

Our key contributions are summarized as follows:
    \textbf{(1)} We introduce a scalable, tuneable, industry-grade framework for data-driven prediction-planning in HD map-free setting. Our model offers an alternative to the e2e model and demonstrates performance on par with off-the-shelf, optimization-based planner.
    \textbf{(2)} To the best of our knowledge, we are the first evaluation of such system in a complex, real-world, urban road with factory ready sensor set and compute platform\footnote{Road test video is available at \url{https://youtu.be/DYPB5lCRvn0}}. We demonstrate the imperative of closed-loop evaluation compared to imitation metric adopted by extensive offline studies.
    \textbf{(3)} We extend the state-of-the-art data-driven approach by proposing a method that goes beyond simple imitation or trajectory sampling. Instead, our method reasons about the potential in generative learning by combining max margin planning with multi-modal imitation, as depicted in \cref{fig:fig0}

\section{Related work}
\subsection{End to End Learning\&Validation}
End-to-end autonomous driving has become a significant research focus. These methods, using neural networks, produce planning trajectories or control commands, with many\cite{shao2022interfuser, wu2022trajectoryguided, mile2022} validations on closed-loop simulator\cite{pmlr-v78-dosovitskiy17a, li2021metadrive}.
However, a persistent domain discrepancy exists between the simulated environment and the actual world, the accurate modeling of agent behaviors and the simulation of real-world data\cite{gao2022nerf} within closed-loop environments continue to pose formidable challenges that have yet to be resolved.
Recently, open-loop experiments with real-world data have attracted more attention\cite{jiang2023vad, hu2023_uniad, hu2022st, zeng2021endtoend}. 
Methods like UniAD\cite{hu2023_uniad} claim their effectiveness in improving final planning performance by involving learning intermediate tasks. 
Nevertheless, it necessitates an auxiliary optimization-based module to generate a viable trajectory and avert collisions, thereby contravening the foundational objective of circumventing deficiencies inherent in optimization-based approaches.
Moreover, the effectiveness of open-loop demonstration has been challenged by \cite{li2023ego} in terms of imbalanced data distribution and metric limitation.
Those unresolved limitations make the results in open-loop or closed-loop simulator less convincing and leaves the effectiveness of methods (like UniAD\cite{hu2023_uniad}, VAD\cite{jiang2023vad}, mile\cite{mile2022}, ST-P3\cite{hu2022st}) unknown for onboard road tests and mass production.
We address these contentious issues by rigorously validating our approach through real-world closed-loop road test, wherein the proposed methodology eschews the dependence on conventional optimization-based algorithms.

\subsection{Data Driven Prediction}
Prior researches have extensively explored actor-based prediction methods for future driving dynamics\cite{chai2019multipath,casas2020importance, phanminh2020covernet, zhao2019multiagent,Rhinehart-2018-109777, tang2019multiple, gupta2018social, lee2017desire}. 
In order to encode scene contexts, a classic CNN architecture can be applied to a rasterized map to predict 2D coordinates \cite{cui2019multimodal}.
More recent methods have started to leverage the topologic graph obtained from HD maps in order to better represent lane connectivity. 
VectorNet \cite{gao2020vectornet} encodes map and agent as polylines with a global interaction graph. LaneGCN \cite{liang2020learning} treats actor histories and the lane graph separately and then fuses them with a series of attention layers.
Temporal occupancy maps \cite{jain2019discrete, ridel2019scene, hong2019rules}, on the other hand, provide another option to forecast grid or heat map at the scene level. We follow the philosophy of \cite{elfes1989using, hoermann2018dynamic, sadat2020perceive, thrun2003learning, wu2020motionnet} in modeling scene-level representations but propose a multi-stage pipeline to perform forecasting in both formats. 
Moreover, unlike typical methods, which heavily rely on the precision of trajectory output, 
our agent results are primarily used to enhance the safety layer and to boost some of our samplers.
\subsection{Data Driven Planning}
Apart from the aformention end to end works, imitation-based methods have been widely explored in the planning task.
SafetyNet\cite{vitelli2022safetynet} developed a machine learning planner and tested it in downtown San Francisco, but failed to avoid collision without the help of a fallback layer. 
VAD\cite{jiang2023vad} modeled the driving scene as a fully vectorized representation.
PlanT\cite{renz2022plant} proposed a transformer-based backbone in actor-level representation and validated in simulation.
However, the monotonous trajectory regression suffers from robustness and lacks interpretability\cite{codevilla2019exploring}, thus difficult to tune by case and hard to scale up.
More recently cost-volume-based methods have improved adaptability and interpretability in challenging environments. They identify optimal trajectories by evaluating candidates within a network-learned cost volume\cite{zeng2021endtoend, sadat2020perceive, casas2021mp3, zeng2020dsdnet, Cui_2021}, while the candidate trajectories are generated through rule-based samplers.
Beginning with pioneering work on NMP \cite{zeng2021endtoend}, DSDNet\cite{zeng2020dsdnet} and P3\cite{sadat2020perceive} made further developments that combined hand-made and learning-based costs.
In the follow-up works, MP3 \cite{casas2021mp3} introduced a mapless approach, while ST-P3\cite{hu2022st} explored vision-based inputs.
We also adopt this combination to choose the best trajectory, but in contrast to all methods above, which solely rely on a rule-based sampler, we explore the potential of multi-modal imitation and generative learning in motion planning task. 

\begin{figure*}[ht]
    \centering
    \includegraphics[width=1\linewidth]{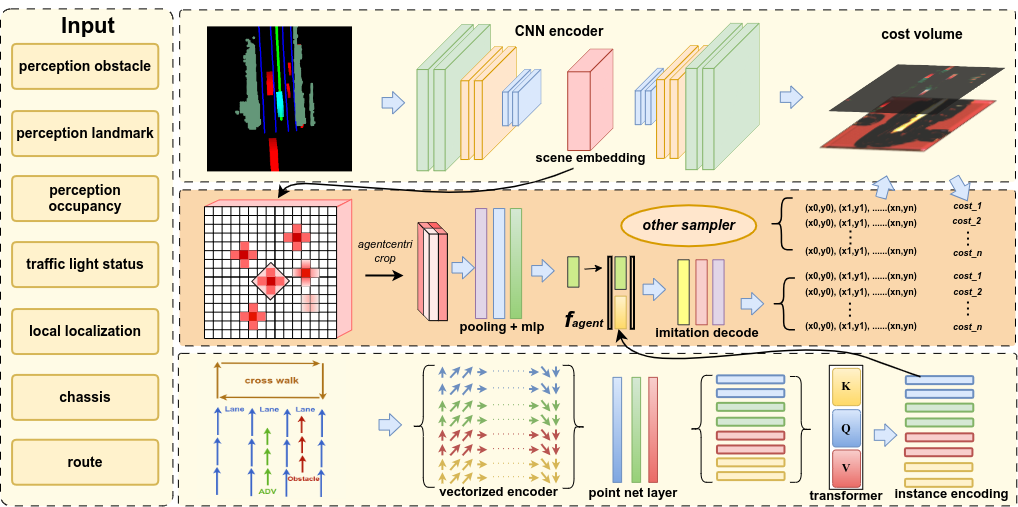}
    \caption{
    The Model Structure of GAD initiates with two streams, rasterization and vectorization, to encode scene and instance-level features; the rasterization branch uses convolution layers to derive the scene embedding, which is decoded into cost volumes and prediction grid maps, while the scene embedding is cropped and combined with the instance encoding from the vectorization branch to form $f_{agent}$ which decoded into ego trajectory and agents' future movements.}
    \label{fig:fig1}
\end{figure*}

\section{Methods}
\subsection{Input Representation}
\label{sec:per}
The goal of this paper is to present a deep learning (DL) pipeline that goes from prediction to motion planning in an integrated manner; hence, it is not our intention to cover perception at length. In our work, the perception module follows a typical mainstream design that provides tracked entities, landmarks, and occupancy voxels (also known as volumn pixels) by incorporating images from a suite of independent cameras with different field of views (FoVs) to formulate a surround view at 360 degrees. 
More details are provided in \cref{app:percep}.
To encode the upstream output for diverse applications, we have devised representations at both the scene and instance levels\footnote{Here, by $\textit{scene}$, we refer to the comprehensive environmental context pertinent to the current frame, and by $\textit{instance}$, we denote the constituents within the scene such as the ego vehicle, other agents, landmarks, routes, etc.}, which are systematically utilized in the evaluation and generation of trajectories.

\paragraph{Scene Representation}
We employ the rasterization methodology as proposed by \cite{djuric2020uncertainty} for the representation of scenes. Rasterization, when coupled with convolutional forwarding, more effectively captures scene-level cues than instance vectorization.
Due to the inherent ability of convolutional networks to exploit spatial correlations, it is feasible to integrate undefined or irregular elements such as construction zones, traffic cones, and roadblocks into a cohesive design, irrespective of their arrangement or geometric regularity. Furthermore, scene information characterized by free space or occupancy, which lacks a parametric framework, is challenging to incorporate within a vectorized representation and, if excluded, poses significant risks to safety-critical systems. 
BEV rasterization assimilates environmental cues from online perception outputs to the fullest extent to capture those 'devils in detail'. 
Comprehensive details are provided in \cref{app:scene}.


\paragraph{Instance Representation}
The instance representation is utilized to mitigate the deficiencies inherent in scene encoding, which suffers from the fact that rasterization tends to distort highly organized and structured data into non-parametric forms. Given that this process is irreversible, it complicates the extraction of instance encoding for elements such as lane centerlines, road boundaries, among scene feature map. On the other hand, vectorized representation combined with the transformer has been proven to be superior to CNN forwarding on BEV raster for instance reasoning\cite{gao2020vectornet}. 
To leverage the modeling capabilities afforded by this form, instances with parametric description are encoded in the vector form, including route points, agent histories, lane center lines, and road boundaries, while non-parametric outputs are ignored, such as occupancy, construction area, roadblocks, and irregular elements. More details can be found in \cref{app:ins}

\subsection{Model Structure}
\label{sec:model_stru}
We build a two-stream backbone network to encode inputs and extract agents and scene features separately. As shown in \cref{fig:fig1}.
One steam processes the rasterized scene while the other deals with the vectorized graph. To elaborate, the rasterization stream employs a convolutional encoder of multiple resolutions to derive the scene embedding, which is then passed through two separate deconvolution decoders to generate the cost volumes and the predicted grid maps respectively.
Meanwhile, in the other stream, inspired by \cite{gao2020vectornet}, the instance-level encoding is built on a hierarchical graph network-based architecture. It consists of a PointNet-based \cite{qi2016pointnet} local sub-graph for processing local information from vectorized inputs and a global graph using a Transformer encoder \cite{vaswani2017attention} for reasoning about interactions over agents and landmark features.
Finally, the corresponding scene encoding for the agent is derived via oriented crop\cite{chai2019multipath} in ego-centric scene feature and fused with instance encoding to obtain agent-level features $f_{agent}$. Then an imitation decoder is utilized to produce the trajectories for agents and the ego.
The details structure for the encoders and decoders are illustrated in \cref{app:struc}.

\subsection{Route}
When HD maps are available, the route is typically a sequence of lanes for the ADV to follow. In settings without HD maps, this approach is not feasible. In this work, route information is derived from visual landmarks and human commuting trajectories for different scenarios to substitute those absent cues. This approach is deemed effective for two primary reasons. Firstly, landmarks (lane center, lane divider and road boundary) provide sufficient cues for driving in a cruise scenario - it generally consists of lane keep and lane change for following and overtaking behaviors regardless of how road topologies are organized. Secondly, for turning scenes(left turn, right turn, u-turn), we gather routing information from historical commuting trajectories, supplemented by real-time occupancy data to mitigate collision risks. The implementation details can be found in \cref{apen:route}.

\subsection{Prediction}
\label{sec:pred}

\begin{figure}[t]
    \centering
    \includegraphics[width=\linewidth]{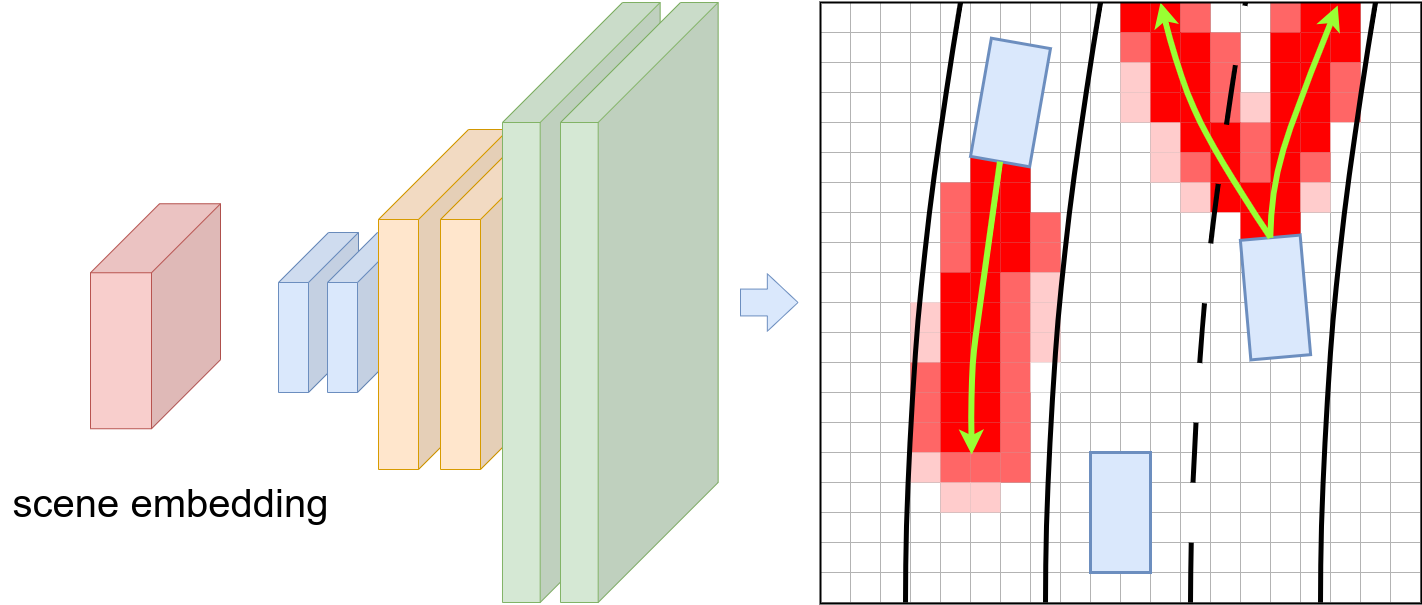}
    \caption{Multi-stage Prediction}
    \label{fig:fig3}
\end{figure}

We employ a multi-stage design for scene and agent prediction as depicted in \cref{fig:fig3}, where the first stage, involves predicting the scene-level grid map (also known as the heat map or occupancy) by regressing $G_{rid} = dec(f_{ras})$ with a resolution that matches the rasterization input through decoding of the scene embeding. Temporal resolution $t$ is 0.1 second from 0 to 3 seconds.

The grid prediction decoder, as mentioned in \cref{sec:model_stru}, shares a similar structure with the cost volume decoder, yet it is supervised from the rasterization of the agents' future ground truth box. This grid is trained with pixel-wise focal loss inspired from \cite{gilles2021home}, for brief we replace $G_{rid}$ with $g$:
\begin{equation} \label{eq:0}
\begin{array}{r}
    L_{pred\_grid} = -\sum_{ijt} [\tilde{g_{ijt}}(1 - \hat{g_{ijt}})^{2} log(\hat{g_{ijt}}) + \\(1-\tilde{g_{ijt}})\hat{g_{ijt}}^{2} log(1 - \hat{g_{ijt}})] 
\end{array}
\end{equation}
Here, $\tilde{g_{ijt}}$ represents the class label for the grid, where $\tilde{g_{ijt}} = 1$ indicates that it is occupied in timestamp t.
In the second stage, the model forecasts agent-level trajectories with the scene prior from the predicted grid by following an interactive design. Since the output trajectories are mainly adopted in sampling for ADV's following and overtaking, it captures only nearby agents, assuming interactions with the ego. In particular, we gather the corresponding probability of nearby landmarks by accumulating grid probability belonging to those areas $P_{lane} = \sum_{t}\sum_{i,j \in(lane)}g_{ij}^{t}$, filtering out higher probability candidates as the target prior. These landmarks indicate likely interaction areas, and agents' trajectories are decoded based on this lane prior:
\begin{equation}\label{eq:pre}
    Traj = dec(f_{agent} | f_{lane})
\end{equation}
Here, $f_{agent}$ is the instance encoding as illustrated in \cref{sec:model_stru}, $f_{lane}$ is the corresponding lane encoding obtained in the vectorized stream.
For the training phase, the ground truth $\tilde{lane}$ corresponding to the agent's future trajectory will be labeled and incorporated as a condition prior:
\begin{equation}
    L_{traj\_pred} = L2(\tilde{Traj}, dec(f_{agent} | f_{\tilde{lane}}))
\end{equation}
The training loss is measured by the $L2$ distance between the predicted and ground truth trajectory.

\subsection{Trajectory Evaluation}
\label{sec:traj_eval}

The previous section introduced a multi-stage prediction method that produces grid maps and interactive trajectories. We now demonstrate the grid can be seamlessly integrated with perception occupancy into an end-to-end motion planner utilizing space-time cost volumes to evaluate candidate trajectories. Following the framework of \cite{zeng2021endtoend} for max-margin planning, we refine it to include scene occupancy and anticipated grid maps, and our evaluator decodes a space-time cost volume $C_{xy}^{t} = dec(f_{ras})$ in grid form for future 30 consecutive frame from BEV raster. The trajectory cost is calculated as the sum of costs at its space-time waypoints, with the best trajectory being the one that minimizes this cost:
\begin{equation} \label{eq:1}
    s = arg\min_{\hat{s} \in S}\sum_{t} [C(\hat{s}) + \alpha O_{cc}(\hat{s}) + \beta G_{rid}(\hat{s})] 
\end{equation}
where $S$ represents the set of planed trajectories, $O_{cc}$ represents perception occupancy, $G_{rid}$ is the predicted grid map, $\alpha$ and $\beta$ are hyperparameters. Since selecting the minimum-cost trajectory within a discrete set is not differentiable, we use the max-margin loss to penalize trajectories that either deviate from the human driving trajectory or are unsafe, and to encourage the human driving trajectory $\tilde{s}$ to have a smaller cost than other trajectories $\hat{s}$ for a given example:
\begin{equation}\label{eq:2}
    L_{cost\_volume} = \max_{\hat{s}}[C(\tilde{s}) - C(\hat{s}) + Diff(\tilde{s}, \hat{s})]
\end{equation}
The $Diff$ metric functions as a compensatory quantifier, meticulously gauging the disparity between human demonstrations and algorithmic candidates. It is designed to assess the efficacy of human driving and the deficiencies of candidates, as delineated by objectives such as planning target, evasion of static obstacles, and regulation of dynamic interactions. This enables the evaluator to attribute commensurate costs to the given trajectories.
In particularly:
\begin{equation}\label{eq:3}
    Diff(\tilde{s}, \hat{s}) = L2(\tilde{s}, \hat{s}) + O(\hat{s}) + G_{rid}(\hat{s})
\end{equation}
where $O$ is the occupancy cost, $G$ is the rest of the prediction costs as defined in \cref{sec:pred}. The imitation loss $L2$ measures the $l2$ distance between trajectory $\hat{s}$ and the ground-truth $\tilde{s}$ for the entire horizon.

\subsection{Trajectory Generation}
\label{sec:trajgen}
Given the learned cost volume, the final trajectory can then be computed by minimizing \cref{eq:1}. Note, however, that this optimization with physical constraints is NP-hard. On one hand, we evaluate and extend previous works on trajectory sampling\cite{zeng2021endtoend, casas2021mp3, sadat2020perceive, hu2022st} by incorporating prior knowledge and structured outputs from perception and prediction. Furthermore, we leverage a GAN-based pipeline during the imitation generator's training to solve the optimization problem. In this section, we describe our sampling and generation algorithm.

The sampling and generating algorithms are categorized into reference line-based and reference line-free types, where methods that incorporate reference information provide long-term consistent outputs and enhance cross-frame stability as this information remains largely unchanged over time. However, challenges arise in scenarios where reference-related information is scarce. To improve our system's scalability, we follow \cite{zeng2021endtoend, casas2021mp3} to develop curve and retrieval-based samplers in a reference-line free context, detailed in \cref{app:curve} and \cref{app:retrival}. Additionally, this section will discuss the proposed lattice sampler, imitation planner, and GAN-based planner, which utilize reference information constructed from visual landmarks or commuting trajectories.

\paragraph{Lattice Sampler}
The lane centers and surrounding agents' movements are strong priors to construct the potential trajectory to be executed by the ADV. When landmarks are available, the lane geometry can be exploited to guide the trajectory sampling process. The predicted trajectory, on the other hand, can be incorporated to generate interactive behavior like following or overtaking in the longitudinal direction. In our paradigm, they are both available as depicted by \cref{sec:pred} \cref{sec:per}. In particular, we follow the trajectory parameterization and sampling philosophy proposed in \cite{werling2010optimal, sadat2019jointly}. The sampled trajectories include maneuvers such as lane keeping, lane changing, nudging, side passing in the lateral direction, as well as following, accelerating, yielding, and stopping in the longitudinal direction. More details can be found in \cref{app:lattice}.

\paragraph{Imitation Planner}
For the imitation planner, previous approaches choose to mimic a human driver with one single trajectory\cite{hu2023_uniad, jiang2023vad}.
But such designs suffer from model collapse in both longitudinal and lateral directions. 
Owing to our generator-evaluator paradigm, we can generate a bunch of diverse candidates and pass them to the downstream evaluation task.
We follow multiple-trajectory\cite{cui2019multimodal} design to estimate the conditional distribution $P(S|f_{plan})$ and multi-modality in motion planning task:
\begin{equation}
    P(S|f_{plan}) = \sum_{M} P(S|m, f_{plan})P(m|f_{plan})
\end{equation}
Where $S=[\hat{s_{1}}, ...\hat{s_{m}}]$ is the trajectories with $m$ different maneuvers such as lane following, lane changing, agent following, agent overtaking. $f_{plan}$ is constructed by feature prepared for planning task in \cref{eq:4}, where ($||$) stands for concatenation.
\begin{equation}\label{eq:4}
    f_{plan} = [f_{scene}||f_{route}||f_{ADV}||f_{target\_speed}]
\end{equation}
The imitation generator decodes $M$ possible trajectories for ego move in local coordinates. During training, we then identify modal closest to the ground-truth trajectory according to $L2$ distance.

\begin{equation}
    {[\hat{s_{1}}, ...\hat{s_{m}}]} = G(f_{plan})
\end{equation}
\begin{equation}\label{eq:6}
    L_{imit\_plan} = arg\min_{i\in{1,...m}}dist(\tilde{s}, \hat{s_{i}})
\end{equation}

\paragraph{GAN-Based Planner}
The aforementioned imitation generator is trained independently from our evaluator under distinct supervisory decoder. In pursuit of generating trajectories utilizing the acquired cost volume and addressing \cref{eq:1} through a learning-based approach, we propose a GAN-based planner with a conditional generative adversarial training framework. This framework is designed to bridges the isolated optimization targets between generator and evaluator, thus facilitating the learning of a stochastic generative model beyond trajectory imitation, adept at encapsulating the multi-modal diversity inherent in planning trajectories. Regarding the standard adversarial loss\cite{goodfellow2020generative}, commonly referred to as the min-max loss, the generator endeavors to minimize the subsequent function, while the discriminator seeks to maximize it.
\begin{equation}
    L_{gan} = \min_{G}\max_{D}(E_{x}[log(D(x))] + E_{z}[log(1-D(G(z)))])
\end{equation}
$log(D(x))$ refers to the probability that the discriminator is rightly classifying the real case, maximizing $log(1-D(G(z)))$ would help it to label the fake sample that comes from the generator correctly.
In our proposed framework, the conventional discriminator is supplanted by the evaluator as delineated in \cref{sec:traj_eval}. Contrary to the traditional discriminator in GANs, which categorizes instances with qualitative labels denoting veracity, our evaluator quantifies the divergence among candidates through the learned cost volume.
Thereby, the optimization target for evaluator \cref{eq:2} still holds, but the learning target of the generator changed. The complete formula derivation is provied in \cref{deduction}. 
\begin{equation}\
    L_{generator} = E(G(m)) + L_{imit\_plan}(\tilde{s}, G(m))
\end{equation}
\begin{equation}
    L_{evaluator} = \max_{\hat{s}}[E(\tilde{s}) - E(\hat{s}) + Diff(\tilde{s}, \hat{s})]
\end{equation}
with:
\begin{equation}\label{eq:E}
    E(s) = \sum C(s) + \alpha O_{cc}(s) + \beta G_{rid}(s) 
\end{equation}
\begin{equation}\label{eq:gang}
    [\hat{s_{1}}, ...\hat{s_{m}}] = \hat{S}= G(m); m \sim \mathcal{N}(\mu,\,\sigma^{2})\,.
\end{equation}
The architecture is exactly the same as presented in previous subsections, except that in the generative model, the hidden variable $m$, which indidates driving manual in deterministic model is sampled from a Gaussian distribution.
The final encoding for ADV is concatenated with $m$ to decode into its trajectory.
However, the trajectory evaluation \cref{eq:E} involve index trajectory points in the learned cost volume $C_{xy}^{t} = dec(f_{ras})$, which is non-differentiable, thus cannot be back propagated to the generator. To address this issue, we use a soft function to get a meaningful gradient. The grid\_sample operator\cite{Ansel_PyTorch_2_Faster_2024} is adopted here to obtain the gradients via interpolation between values in our cost volume.

\subsection{Safety layer}
\label{safety_layer}
For on-road autonomous driving, a strategy for safety coverage is necessary. The perception occupancy and trajectory prediction results are used to validate the trajectory candidates with respect to safety. 
The trajectory prepared for the safety layer is a little bit different to \cref{eq:pre} in that no prior is assumed in decoding.
We sort the trajectories according to the evaluated costs and iterate them to check the basic rule in modularized form(like trajectory and bounding box) until we cache enough candidates by ascending order.
To ensure cross-frame consistency, we reserve $n$ validated trajectories with the lowest cost and compare them with planning output in the last frame.
The above objective promotes the system's robustness with respect to the corner cases that our model fails to handle. We believe such a layer of safety guarantee is indispensable for manufacturing our system at scale.


\section{Experiments}
We develop various baseline variants of our models under both open-loop and closed-loop experiments for comprehensive ablative analyses, and we quantitatively benchmark these against cutting-edge alternative methodologies.  We gather in-house datasets for both training and open-loop evaluations, encompassing an extensive array of driving scenarios within densely populated urban settings. To substantiate the feasibility of the data-driven planners, we deactivate the functionalities in the safety layer as well as the optimizer and safety backup subsystems in comparative methodologies such as optimal control in UniAD\cite{hu2023_uniad}, etc.

\subsection{Prediction}
\begin{table}[t]
  \begin{adjustbox}{max width=\linewidth}
  \centering
  \begin{tabular}{c|ccc|c}
    \toprule
    Method &  \multicolumn{3}{c}{Mean L2 / hit-rate $<$ 1m} & Avg. \\
    {}   & 1s   & 2s   & 3s & {} \\
    \midrule
    VectorNet\cite{gao2020vectornet}      &  \textbf{0.75}/\textbf{0.80}  & 1.72/0.40  & 2.85/\textbf{0.22} & \textbf{1.34}\\
    GAD-uni    &  0.76/0.72  & \textbf{1.70}/\textbf{0.38}  & \textbf{2.83}/0.23 & 1.35\\
    \midrule
    TNT\cite{zhao2021tnt}     &  0.65/0.78  & 1.01/0.57  & 1.72/0.34 & 0.90\\
    GAD-multi  &  \textbf{0.60}/\textbf{0.84}  & \textbf{0.95}/\textbf{0.65}  &\textbf{ 1.59}/\textbf{0.39} & \textbf{0.81}\\
    \bottomrule
  \end{tabular}
  \end{adjustbox}
  \caption{Prediction Results in uni-modal and multi-modal.}
  \label{tab:t1}
\end{table}
For the prediction task, prior state-of-the-art methodologies are appraised within HD map-free context, wherein solely the observable landmarks are furnished to the predictors.
We measure the performance of methods on root mean squared error and hit-rate overall future timesteps $t\in{1, 2, 3}$ seconds.
In \cref{tab:t1}, we compare all methods using the same inputs.
In general, our multi-modal design with interactive prior outperformed our reimplemented TNT\cite{zhao2021tnt} with 16 candidates, while the uni-modal version shared similar results with the current state-of-the-art.


\subsection{Open-loop Planning}
For open-loop planning, our analysis prioritizes two pivotal evaluation metrics: the $L2$ error and the collision rate. The planning horizon is standardized at 3.0 seconds to ensure equitable comparison. We ascertain the $L2$ error by quantifying the deviation of the planned trajectory from the human driving trajectory and assess the frequency of potential collisions involving the ADV with other road agents. A comprehensive comparison with antecedent methodologies is delineated in \cref{tab:t2}. It is noteworthy that the imitation strategy incurs penalties proportional to the trajectory's divergence from the empirical ground truth, resulting in minimal $L2$ errors across all examined horizons.

\begin{table}[t]
    \centering
    \begin{tabular}{c|c|c|c|c|c|c}
    \toprule
        Method & \multicolumn{2}{c}{hist} & tar v & L2 & \multicolumn{2}{c}{passing rate} \\ 
        {} & ras & vec & m/s & Ave & straight & turn \\
        \midrule
        GAD-multi\_1 & $\surd$ & $\surd$ & \large x & 0.56 & 0.00 & 0.00\\
        GAD-multi\_2 & $\surd$ & $\surd$ & $\surd$  & 0.63 & 0.80 & 0.75\\
        GAD-multi\_3 & $\surd$ & \large x & $\surd$ & 0.70 & 0.85 & 0.70\\
        GAD-multi\_4 & \large x & \large x & $\surd$  & \textbf{0.34} & 0.65 & 0.00\\
    \bottomrule
    \end{tabular}
    \caption{Ablation study on ego feature for imitation planner. }
    \label{tab:history}
\end{table}

\begin{table*}[h]
\centering
\begin{adjustbox}{max width=4.5in}
  \begin{tabular}{c|ccccc|ccc|ccc}
    \toprule
    Method &  \multicolumn{5}{c}{trajectory sources} &\multicolumn{3}{c}{L2(m)} & \multicolumn{3}{c}{Collision(\%)} \\
    {}             & curve & retrieval & lane & lattice & model & 1s   & 2s   & 3s   & 1s  & 2s  & 3s \\
    \midrule
    UniAD\cite{hu2023_uniad}       &{}   &{}   &{}   &{}  &$\surd_{uni}$  &0.14   &0.44   &0.92   &0.17  &0.33  &0.78 \\
    VAD\cite{jiang2023vad}         &{}   &{}   &{}   &{}  &$\surd_{uni}$  &\textbf{0.12}  &\textbf{0.41}   &\textbf{0.89}   &0.15  &0.30  &0.69 \\
    \midrule
    NMP\cite{zeng2021endtoend}         &$\surd$   &{}  &{}   &{}   &{}  &0.25   &1.19   &2.96   &\textbf{0.08}  & \textbf{0.16}  & 1.05 \\
    MP3\cite{casas2021mp3}         &{}   &$\surd$  &{}   &{}   &{}  &0.42   &1.31   &2.73   &0.09  & 0.22  & 0.97 \\
    P3\cite{sadat2020perceive}          &{}   &{}  &$\surd$   &{}   &{}  &0.24   &1.55   &3.95   &0.14  & 0.35  & 1.04 \\
    \midrule
    GAD-multi  &{}  &{}  &{}  &{}  &$\surd_{multi}$   &0.16   &0.69   &1.26   &0.11  & 0.31  & 0.52 \\
    GAD-ref\_free  &$\surd$   &$\surd$ &{}  &{}   &{}  &0.25   &0.90   &2.15   &0.09  & 0.17  & \textbf{0.48} \\
    GAD-base        &{}   &{}   &{}   &$\surd$   &$\surd_{multi}$  &0.20   &0.84   &1.84   &0.10  & 0.31  & 1.07 \\
    GAD-gan        &{}   &{}   &{}   &$\surd$   &$\surd_{gan}$  &0.18   &0.80   &1.67   &0.09  & 0.28  & 1.02 \\
    \bottomrule
  \end{tabular}
  \end{adjustbox}
  \captionsetup{width=4.5in}
  \caption{Open loop planning results of imitation-based method, sampling-based method and GAD.}
  \label{tab:t2}
\end{table*}

\subsection{Ablation of Ego Feature}
A potentially misleading aspect of current data-driven methodologies in motion planning may be the excessive reliance in utilization of ego history. Historical way points are pivotal in predictive tasks for inferring the future trajectories of agents. Conversely, planning tasks diverge in that the trajectory is synthesized to achieve specific longitudinal and lateral objectives such as velocity and lateral position, thereby necessitating minimal dependence on historical motion.
We conduct experiments to validate the necessity of ego history and target speed encoding in \cref{tab:history}. The historical information is derived from two distinct sources within our framework: one pertaining to rasterized ego motion and the other to vectorized historical positions. Since lateral targets can be deduced from route and landmarks, we only provides feature encoding in target speed of 3s-th to guide trajectory generation(it can be conceptualized as a data-driven sampler analogous to the cruise scenario strategy within lattice framework but only concerns the speed in 3s-th). Such setting provides a tunable parameter in deployment for the users to adjust the desired cruise speed range. As show in the \cref{tab:history}, we evaluate the open-loop metric and the closed-loop scene passing rate of lane keeping and turning for imitation planner. The results indicate that history feature dose help to improve the open-loop metric, but target speed feature is the key for success closed-loop. Method without history encoding failed in turning scene.

\subsection{Closed-loop Experiments}

Initially, we conducted an ablation study to assess the necessity of various samplers within a closed road environment. Subsequently, we identified the optimal sampler combination for further evaluation on open urban roadways. 
Comprehensive real-world testing of these methodologies was conducted in densely populated urban settings, under the vigilant supervision of human safety drivers. The specific driving route utilized for our experiments is depicted in \cref{roadtest}. Throughout the extensive public road testing, spanning over 30 kilometers, the model adeptly executed a diverse array of complex maneuvers, such as lane-following, merging, yielding to pedestrians. The results of these road tests are documented in \cref{tab:t3}.
\begin{table*}[h]
  \begin{adjustbox}{max width=\textwidth}
  \centering
  \begin{tabular}{c|ccc|cccc|c|c}
    \toprule
    Method &  \multicolumn{3}{c}{comfort}&\multicolumn{4}{c}{event in 30Km} & efficiency & MPI \\
    {} & Lat. accel. & Jerk & $\Delta$Steering Angle & Collisions   &  Lost control  &  Discomfort &  Deviation & Minute & {}\\
    \midrule
    UniAD\cite{hu2023_uniad}     &0.351   &3.12   &3.68   &-  & - & - & - & 15.20 & 0.31  \\
    \midrule
    ST-P3\cite{hu2022st}     &0.857 &4.54 &8.53   &4  & 3 & 14 & 3 & 8.62 & 1.36  \\
    \midrule
    GAD-multi    &\textbf{0.489}  &1.22  &2.89  &4  & 1 & 2 & 4 & 5.60 & 2.72\\
    GAD-lattice  &0.728  &2.26  &4.22  &2  & 0 & 4 & 2 & 5.68 & 3.75\\
    GAD-base     &0.531  &1.20  &4.21  &3  & 1 & 3 & 1 & 5.56 & 3.75\\
    GAD-gan      &0.511  &\textbf{1.15}  &3.52  &3  & 1 & 2 & 1 & \textbf{5.51} & 4.28\\
    GAD-gan\dag   &0.492  &1.16  &\textbf{2.77}  &0  & 1 & 2 & 1 & 5.54 & \textbf{7.50}\\
    \midrule
    DPQP         &0.284 &0.85 &2.11   &1  & 0 & 1 & 1 & 5.42 & 10.0  \\
    \bottomrule
  \end{tabular}
  \end{adjustbox}
  \caption{Road test data were systematically collected during a 30 km road test, encompassing the computation of average time per loop. Concurrently, measurements of lateral acceleration, jerk, and steering angle were meticulously recorded. (\dag) indicates boosted by safety layer and cross frame align. The UniAD consistently deviates and collides with obstacles, thereby failing to accumulate sufficient statistical data.}
  \label{tab:t3}
\end{table*}

It is noteworthy that methodologies exhibiting superior imitation metrics do not invariably demonstrate enhanced performance in closed-loop evaluations. This discrepancy may be attributable to the limitations of the imitation metric in adequately assessing the complexities of planning tasks(we illustrated the detail in \cref{app:limitation}). Nonetheless, the multi-modal imitation generator and GAN-based generator significantly enhance the system's performance in both open-loop and closed-loop settings, primarily in terms of the planner's intelligence and the continuity between consecutive frames. The efficiency and comfort metrics both show improvements with the integration of data-driven generators. Closed-loop results suggest that multimodality mitigates the brittleness inherent in imitation learning, while the lattice sampler further enhances the robustness of the entire system. See \cref{app:multi}

To rigorously evaluate the efficacy of the proposed methodology as a substitute for off-the-shelf planners, we conducted a comparative analysis between the GAD and a conventional modularized, user-level, rule-based system, referred to as DPQP in \cref{tab:t3}. Notwithstanding, a discernible performance discrepancy persists in terms of system's robustness and comfortableness. We posit that increase training data size could ameliorate this disparity. The system obtains compariable results with DPQP after boosted by the safety layer and cross frame align as mentioned in \cref{safety_layer}.

\section{Conclusions}
\label{sec:conc}
We introduce a sophisticated DL pipeline from prediction to motion planning in an HD map-free setting, augmented with enhanced safety assurance, which has shown substantial enhancements over prevailing data-driven methodologies in Shanghai's complex urban environments. This paradigm enables the safe deployment of learned planners in real-world scenarios, offers interpretability, and leverages their capacity to evolve with accumulating data, managing more complex scenarios than traditional rule-based systems. 
Future work may integrate generative learning techniques such as CVAE\cite{kingma2013auto} and diffusion models\cite{rombach2022high} to further augment these capabilities. Additionally, our results suggest that while imitation learning can enhance motion planning tasks, the imitation metric alone is insufficient as a definitive criterion for assessing the efficacy of motion planners; thus, it is recommended that subsequent research incorporate this metric to gauge the intelligence domain of systems and develop additional metrics for closed-loop evaluation.

\section{Acknowledgements}
We would like to thank the following contributors who work on the toolchains to support our closed-loop evaluation and onboard road test: Chi.Zhang, Yue.Li, Bolin.Zhao, Yanze.Liu, Liang.Zhu.

{\small
\bibliographystyle{ieee_fullname}
\bibliography{egbib}
}


\appendix
\section{Appendix}
\subsection{Perception Outputs}
\label{app:percep}
At each timestamp, we have 3D detection results for surround view camera, these inputs are fed into an existing tracking system, outputting stat estimates S for all surrounding actors (state encompasses the bounding box dimensions, spatial coordinates, linear velocity, linear acceleration, directional orientation, and the rate of change of heading). Denote a set of discrete times at which tracker outputs state estimates as $[t_{1}, t_{2},...t_{T}]$, where time gap between consecutive time steps is constant (the gap is equal to 0.1s for tracker running at a frequency of 10Hz). 
Furthermore, the perception module also provides information about all perceivable landmarks and road boundaries ahead, measured by the polynomial curve. 
Stop lines are also essential when the ADV approaches the junction area. 
The scene occupancy, on the other hand, is constructed using a grid map with a resolution of 0.2m, where each grid cell is labeled according to its motion status(static or movable). 

We elucidate the rationale behind our architectural that segregates perception from the comprehensive end-to-end stack: Firstly, for mass production friendly concern, the modularized perception provides explaninablity and tunablitiy. Secondly, information articulated in parametric form facilitates reasoning at the agent level, not solely for application within the safety layer and the analysis of explainable cases, but also to enhance our trajectory generation processes. Thirdly, parameterization encompasses post-tracking and fusion for the results of single or multi-modal detection on raw sensor data which remain formidable unresolved issues, particularly when considering the substitution of these processes with a model to attain comparable results. We leave perception involved version to follow up works when those issue resolved in the future.

\subsection{Scene Rasterization}
\label{app:scene}
The primary function of scene rasterization is to map scene information to different channels, enabling the neural network to extract scene semantic information more effectively. For example, obstacles, routes, and lane lines are mapped into the RGB channels, respectively. Specifically, the states of landmarks and road-boundary are rasterized in the same channel, facilitating traffic flow reasoning at intersections.
To capture the past motions of all traffic actors, their bounding boxes at consecutive time steps are rasterized on top of map vector layers.
Each historical actor polygon is rasterized in the same channel as the current polygon but with a reduced brightness level, resulting in the diminishing visual impact over time. 
Brightness level at $t$ is equal to $\max[0, (T-t) * \alpha]$, $T$ is timestamp for current frame, where we set $\alpha = 0.05$ (the example shown in \cref{fig:fig1}).
To distinct ADV from other agents, we rasterize ego's polygon in an isolated channel, as well as its fading history.
The non-moving obstacles, together with occupancy grids of static label, are rasterized in a different channel since they share similar semantic meaning.
The rasterized map contains a rendering of traffic light permissibility in junctions, but we tackle it a little bit trick to handle the robustness.
We render permitted (green light), yield (unprotected), or prohibited (red light ) by masking stop line with a static polygon to exhibit the permissibility of current ego route.
The route information is depicted in an isolated channel illustrated in \cref{apen:route}.
All in all, we obtain a 3D tensor of size $(H, W, C)$, with $H=500, W=500, C=5$  for the map.

\subsection{Instance Vectorization}
\label{app:ins}
Parametric information for certain instance such as trajectories, lanes, and routes are presented in the form of discrete points. In order to enable long-distance interaction, we process these points using a vectorization method to generate a set of instances. For each instance $\mathbf{I}_{i}$, it is comprised of a collection of vectors that belong to the same group:
\begin{equation}
    \mathbf{I}_{i} = [\mathbf{v}_{1}, \mathbf{v}_{2}, ... , \mathbf{v}_{n}]
\end{equation}
As for each vector ${\mathbf{v}_{\textit{i}}}$, it can be defined as:
\begin{equation}
    \mathbf{v}_{i} = [\mathbf{p}_{i}, \mathbf{p}_{i+1}, a_{i}]
\end{equation}
where  $\mathbf{p}_{i}$ represents the $i$th point coordinates, and $a_{i}$ denotes an attribute that indicates its semantic label such as center lines and visible landmarks.

Similar to the process of scene rasterization, stop lines are considered as static agents and can be vectorized using unchanged vectors. Each vector can be seen as a node in a graph. Subsequently, a sequence of nodes belonging to the same category will be combined to form an instance, which will then be interconnected within the graph.

\subsection{Coordinate}
The above mentioned representations are built upon Ego-ENU coordinate system, rather than the fixed-oriented coordinate system\cite{DBLP:journals/corr/abs-1808-05819}, which is derived by rotating the Ego-ENU coordinate by ADV's heading. The Ego-ENU coordinate system is centered around the ADV with the x-axis pointing to the east, the y-axis pointing to the north, and the z-axis pointing upwards. In the Ego-ENU coordinate system, the ADV's historical and future trajectories can be diversely distributed depending on the heading directions, with both x and y coordinates of the same order. This allows for a more balanced representation of the vehicle's motions in both longitudinal and lateral directions.

Moreover, the fixed-oriented coordinate system tends to extend along the x direction while fluctuating in a small range around 0 along the y direction, leading to challenges in accurately modeling the lateral behaviour such as overtaking and nudging. The scale differences between x and y directions causes lateral errors disproportionately large in comparison to longitudinal behaviours.

Here we represent our comparative experiments to illustrate it. We have conducted three experiments to test performances in Ego-ENU and the fixed-oriented coordinate system respectively. To tackle the unbalanced distribution of trajectories in fixed-oriented one, we present the original and one with feature normalization.

\begin{table}[ht]
    \centering
    \begin{tabular}{c|c|c}
    \toprule
        Coordiante system & with normalization & L2(m)\\ 
        \midrule
        Ego-ENU & \large x & \textbf{0.39} \\
        \midrule
        Fixed-Oriented & \large x & 0.95 \\
        Fixed-Oriented & $\surd$ & 0.63 \\
    \bottomrule
    \end{tabular}
    \caption{Comparative experiments of coordinate systems. }
    \label{tab:t5}
\end{table}

\begin{figure*}[h]
    \centering
    \subfloat[\label{fig:gnn_enu}]{\includegraphics[scale=0.3]{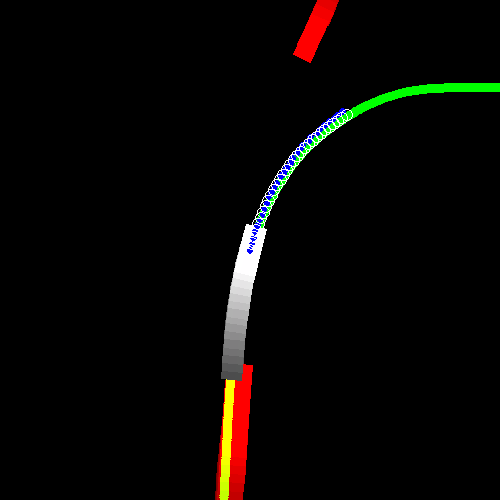}}\hspace{0.02\textwidth}
    \subfloat[\label{fig:gnn_sl}]{\includegraphics[scale=0.3]{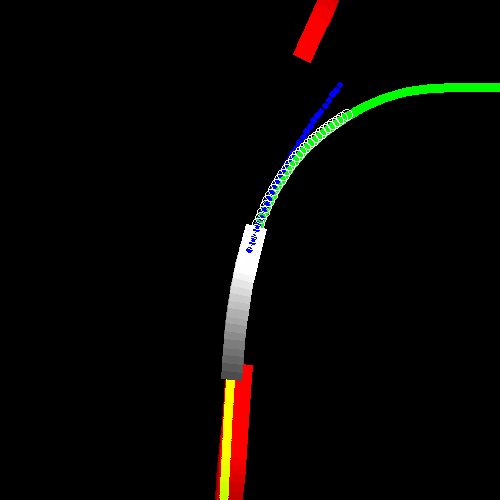}}\hspace{0.02\textwidth}
    \subfloat[\label{fig:gnn_sl_with_norm}]{\includegraphics[scale=0.3]{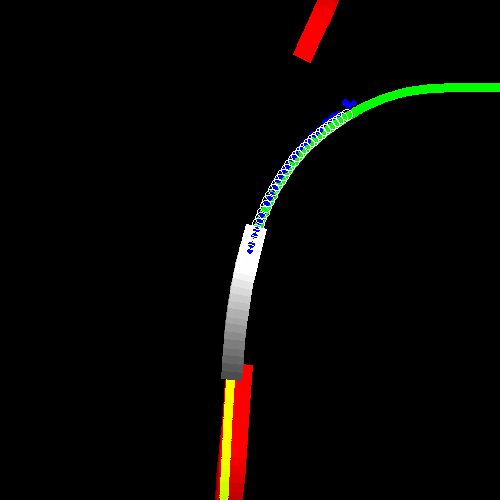}}\hspace{0.02\textwidth}
    \caption{Visualization of comparative results. (a)the predicted trajectory in Ego-ENU closely follows the route. (b)trajectory in the fixed-oriented coordinate system deviates outwards. (c)trajectory in the fixed-oriented coordinate system with feature normalization exhibits proximity to the route.}
    \label{fig:fig4}
\end{figure*}

According to the test results, it is evident that the L2 distance error of the Ego-ENU coordinate system is the lowest since the predicted future trajectory (noted in blue) approximates to the ground truth(noted in white) and closely follows the route(noted in green). Meanwhile, the trajectory in the fixed-oriented coordinate system with feature normalization, also exhibits proximity to the route. However, its longitudinal points are irregularly and non-uniformly distributed. This irregularity is attributed to the scaling of the longitudinal features to the similar order as the lateral ones. Conversely, the trajectory in the fixed-oriented coordinate system without any form of normalization demonstrates the poorest performance, as the end of the trajectory fails to align with the route and deviates outwards, as shown in \cref{fig:fig4}.

Therefore, the more symmetrical Ego-ENU coordinate system, which better represents the ADV trajectories, allows for better understanding of vehicle behaviours and enables more effective trajectory planning and overall autonomous driving performance.

\subsection{Model structure}
\label{app:struc}
Our model architecture consists of two main branches. The first rasterization branch, utilizes four convolutional blocks with varying scales to extract short and long-range features, which are then used to derive the scene embedding. Subsequently, the output from the rasterization branch is fed into two decoders with similar structures, each containing four sub-blocks of deconvolution layers, allowing for step-by-step upsampling to match the size of the input from the rasterization branch. The decoders generate cost volumes and prediction grid maps.

Concurrently, the vectorization branch processes parametric instance information through a PointNet subgraph, which includes three edge convolutions. The output from this branch is then passed through a downstream Transformer encoder to obtain the instance encoding. Note that the rasterization branch has more layers than the vectorization branch as the input is much denser than the vectorized graph. 

Following the extraction of the scene embedding, features by oriented crop around the agents are obtained. These features are then processed through a sequence of max-pooling and a multi-layer perceptron (MLP). The resulting features are concatenated with the instance encoding to create the agent encoding, which is subsequently decoded by an MLP to obtain each agent's trajectory.

\subsection{Route}
\label{apen:route}
\begin{figure*}[h]
    \centering
    \subfloat[\label{fig:a}]{
		\includegraphics[scale=0.4]{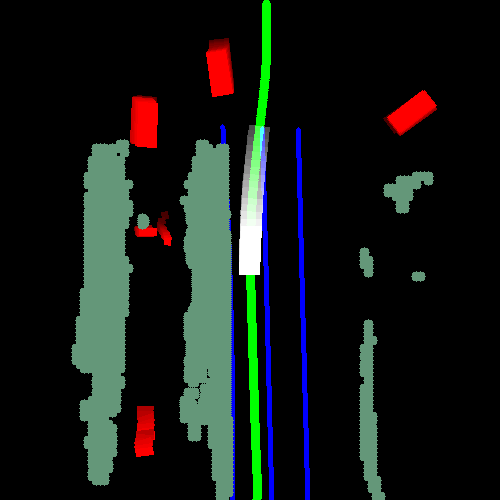}}\hspace{0.02\textwidth}
    \subfloat[\label{fig:b}]{
		\includegraphics[scale=0.4]{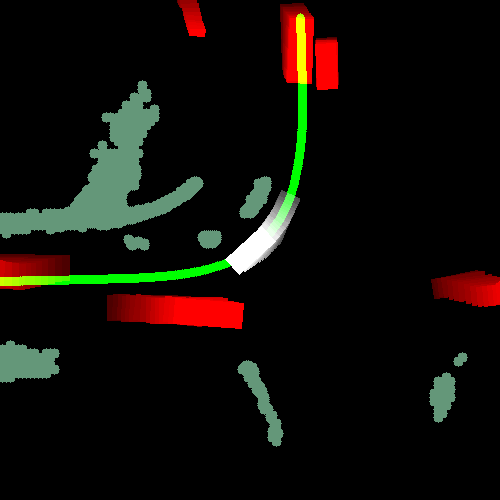}}
    \caption{Route visualization. (a) lane center line is served as route in cruising scenarios. (b) is the  turning scenarios, where historical commuting trajectories are served as route, and occupancy provides road boundary information.}
    \label{fig:route_vis}
\end{figure*}

As indicated previously, the route plays a critical role in guiding ADVs along roadways. Our route information comes from two primary channels: the lane center line and historical commuting trajectories. As shown in \cref{fig:route_vis}. 

The lane center lines, constructed by landmarks which can be accessed in perception outputs, are mainly utilized for cruising scenarios, as the ADV consistently travels along the middle of the lane. We construct three reference lines base on lane center points from current, left and right landmarks to support lane keeping and changing.


However, in turning scenarios, the absence of landmarks without HD-maps results in the ADV lacking appropriate instructions to navigate intersections. To address this, the starting and ending positions from commuting trajectory are recorded. As the ADV approaches intersections, the history human-driving trajectories are employed as the route to guide the ADV through the intersections. Owing to the absence of the road boundary, the scene occupancy has been adopted to delineate the boundary, as illustrated in \cref{fig:route_vis}

Note that in the following visulizations, we omit the occupancy channel for brief.

\subsection{Multi-modal Trajectory Generation}
\label{app:multi}
\begin{figure*}[h]
    \centering
    \subfloat[\label{fig:mm_lc_1}]{\includegraphics[scale=0.3]{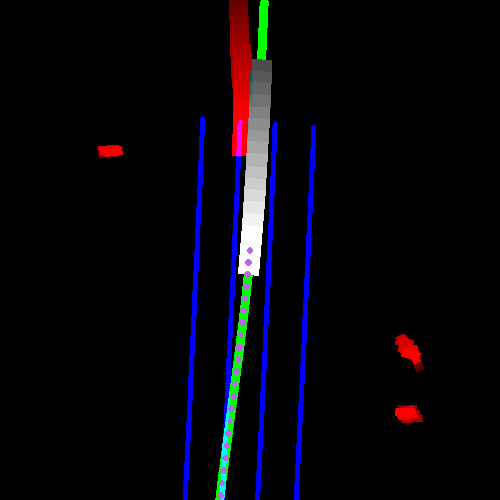}}\hspace{0.02\textwidth}
    \subfloat[\label{fig:mm_lc_16_1}]{\includegraphics[scale=0.3]{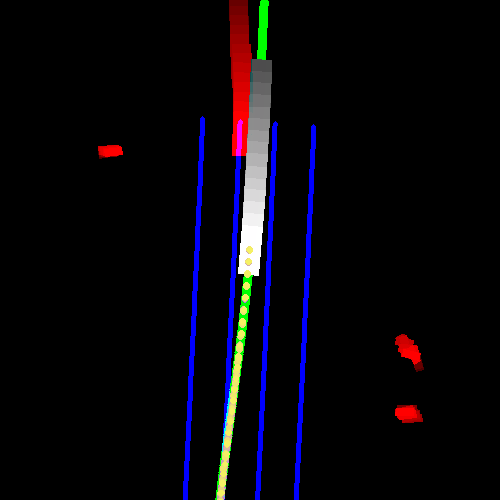}}\hspace{0.02\textwidth}
    \subfloat[\label{fig:mm_lc_16_opti}]{\includegraphics[scale=0.3]{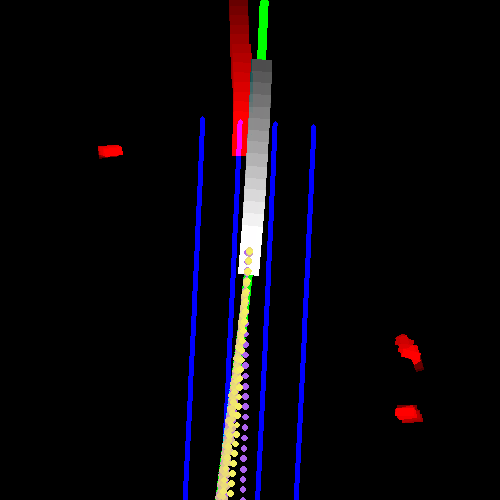}}\hspace{0.02\textwidth}
    \caption{Lane changing scenarios with multi trajectory generation. (a)single trajectory; (b)multi trajectories imitation; (3)GAN-based generator}
    \label{fig:right_merge}
\end{figure*}

\begin{figure*}[h]
    \centering
    \subfloat[\label{fig:mmt1}]{\includegraphics[scale=0.3]{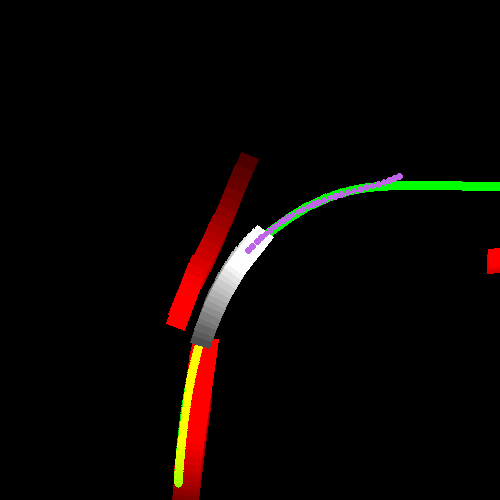}}\hspace{0.02\textwidth}
    \subfloat[\label{fig:mmt2}]{\includegraphics[scale=0.3]{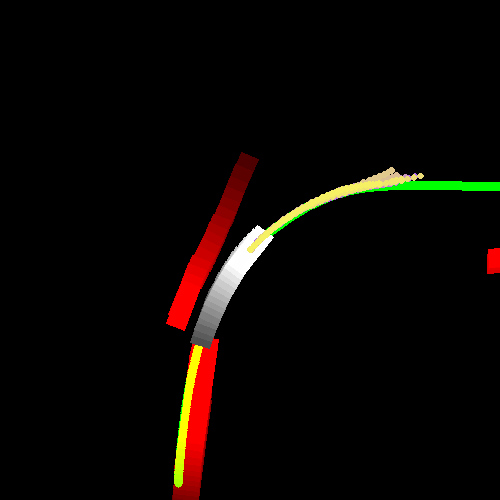}}\hspace{0.02\textwidth}
    \subfloat[\label{fig:mmt3}]{\includegraphics[scale=0.3]{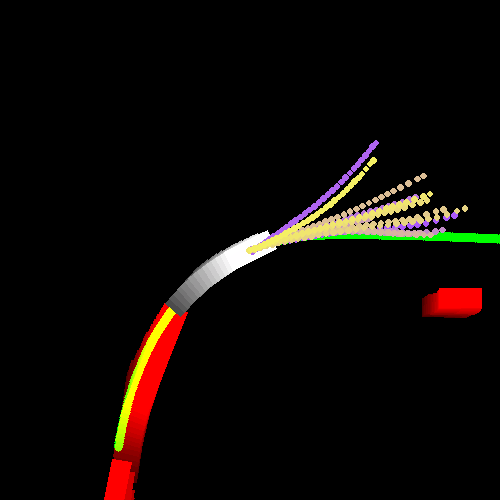}}\hspace{0.02\textwidth}
    \caption{Right turning scenarios with multi trajectory generation. (a)single trajectory; (b)multi trajectories imitation; (3)GAN-based generator}
    \label{fig:mmt}
\end{figure*}

We provide qualitive result of single modal, multi-modal and GAD-based generator in \cref{fig:right_merge} and \cref{fig:mmt}.
GAN-based generator help boosting the multi modality in trajectory generation.

\subsection{Curve-based Sampler}
\label{app:curve}
We follow \cite{zeng2021endtoend} with little modifications for comfort concerns. 
Vehicle trajectories can be categorized into two primary types: straight and turning. Turning trajectories encompass lane changes and turning at intersections. Lane-changing trajectories are characterized by variable curvatures over time, while turning at intersections follows circular-like curves. As straight lines can be treated as a special case of curves, we introduce the curve-based sampler that focuses on generating straight, constant curvature circle-like, and variable curvature clothoid-like trajectories.

For straight ones, a sequence of acceleration values are sampled to account for various velocity-changing scenarios, which are subsequently integrated over time to obtain speed profiles. With initial states(the position and heading angle) of the autonomous vehicle, a set of straight trajectories can be obtained.

Circle trajectories employ the bicycle model\cite{bicyclemodelpaper} to establish the relationship between steering angle $\phi$ and path curvature $\kappa$, represented by the formula $\kappa =2tan(\phi)/L\approx 2\phi/L$, where $L$ is the distance between the front and rear axles of the autonomous vehicle. Using a constant curvature and different sampled speed profiles, a set of circle trajectories can be derived.

Clothoid curves involve a proportional relationship between the curvature $\kappa$ of a point and its distance to the original point $\xi$, which can be defined as

\begin{equation}
    \mathbf{s(\xi)} = \mathbf{s_0} + a \left[ C\left (\frac{\xi}{a}\right ) \mathbf{T_0}  + S\left (\frac{\xi}{a}\right ) \mathbf{N_0} \right] \tag{1}
\end{equation}

\begin{equation}
    C(\xi) = \int_{0}^{\xi} {\rm cos} \left( \frac{\pi u^2}{2} \right)du \tag{2}
\end{equation}

\begin{equation}
    S(\xi) = \int_{0}^{\xi} {\rm sin} \left( \frac{\pi u^2}{2} \right)du \tag{3}
\end{equation}

Where $\mathbf{s(\xi)}$ is the clothoid curve, $\mathbf{T_0}$ and $\mathbf{N_0}$ are the tangent and normal vector of point $\mathbf{s_0}$, $S(\xi)$ and $C(\xi)$ are associated coefficients and defined as the Fresnel integral. Given the initial curvature obtained from the bicycle model and the speed profiles, we can generate a variety of clothoid trajectories by considering a suitable range for the scaling factor $a$.

\subsection{Retrieval-based Sampler}
\label{app:retrival}

The retrieval-based sampler does not depend on the map or lane information to generate trajectories, instead, it retrieves a cluster of trajectories whose initial state are close to the ADV’s current state from a pre-established expert trajectory dataset.\par
To build the expert trajectory dataset, we utilized about 200,000 driving trajectories from experienced human drivers. However, a substantial portion of the original human driving trajectories contains duplicate or very similar samples. Drawing inspiration from \cite{casas2021mp3}, we conducted subdivision and clustering on the original human driving trajectories in order to preserve diversity while simplifying the expert trajectory dataset.\par
To subdivide the original trajectories into different bins, we use the initial velocity \textit{v}, acceleration \textit{a} and curvature \textit{$\kappa$}, with respective bin sizes of $1.0 \frac{m}{s}$, $0.5 \frac{m}{s^{2}}$, $0.01 \frac{1}{m}$. In each bin, trajectories are clustered into 2000 sets, and the closest trajectory to each cluster center is retained. Finally, the trajectories from each subset will be aggregated to the comprehensive expert trajectory dataset.\par
Furthermore, we use kd-tree to organize the expert trajectory dataset for speeding up the query and sampling of ADV. Each trajectory is constructed as a node in the k-d tree, with its initial state ( $\textit{v}_0$, $\textit{a}_0$, $\textit{$\kappa$}_0$ ) serving as the basis for the tree's partitioning. \par

When it comes to sampling, the current state of ADV will be used as a query vector \textbf{\textit{q}}=($\textit{v}_{ADV}$, $\textit{a}_{ADV}$, $\textit{$\kappa$}_{ADV}$) to select trajectories from the k-d tree and the distance function for k-d tree nodes is defined as:
\begin{equation}
\textit{D} = \beta_1 \cdot |v - v_{ADV}| + \beta_2 \cdot |a - a_{ADV}|+\beta_3 \cdot |\kappa - \kappa_{ADV}| \tag{1}
\end{equation} \par
In this paper, we assign different weights to the differences in velocity \textit{v}, acceleration \textit{a} and curvature \textit{$\kappa$}, with $\beta_1=1.0$, $\beta_2=5.0$, $\beta_3=40.0$ and all nodes with $D < 1.0$ will be added into output for the retrieval-based sampler.

\subsection{Lattice Sampler}
\label{app:lattice}
The lattice sampler is a reference line-based method used to acquire multiple candidate trajectories along the reference line, taking into account future movements of surrounding obstacles to avoid potential hazards.

To begin, this method involves sampling multiple candidate trajectories based on the Frenet frame, where the smooth lane central line is projected to the S coordinate and its normal direction to the L coordinate. Trajectories are composed of longitudinal and lateral paths, with the longitudinal one described as the station and time, denoted as $S=\left\{ s(t) \right\}$, and the lateral path described as the lateral station and longitudinal station, denoted as $L={l(s)}$. 

When sampling trajectories, given the initial state $[s_0, s_0^{'}, s_0^{''}], [l_0, l_0^{'}, l_0^{''}]$ at time $t_0$, the end states $[s_1, s_1^{'}, s_1^{''}], [l_1, l_1^{'}, l_1^{''}]$ at time $t_1$ are sampled to cover cruising, following, overtaking and stopping maneuvers, and consequently fits polynomials.

\begin{figure}[h]
    \centering
    \includegraphics[width=0.5\linewidth]{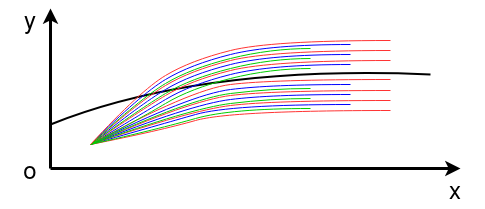}
    \caption{Sampling in the Frenet frame, and then converted back to the Cartesian coordinate system.}
    \label{fig:lattice sampling}
\end{figure}

As for cruising maneuvers, vehicles must adhere to a specific cruising speed by adjusting their acceleration and deceleration. Therefore a set of prediction time horizon $t_1$ and end-condition longitudinal velocities $s_1^{'}$ are uniformly sampled within a pre-defined range of the cruising speed. This sampling process is facilitated by a quartic polynomial.

When it comes to stopping maneuvers, vehicles are obliged to come to a complete stop within a designated distance in order to comply with traffic signals or ensure a safe distance from stationary obstacles. In this scenario, the end states such as speed and accelerations are fixed at 0, while the prediction time $t_1$ and end longitudinal distance $s_1$ are sampled. These parameters are then described using a quintic polynomial.

\begin{figure}[h]
    \centering
    \subfloat[\label{fig:st-a}]{
		\includegraphics[scale=0.3]{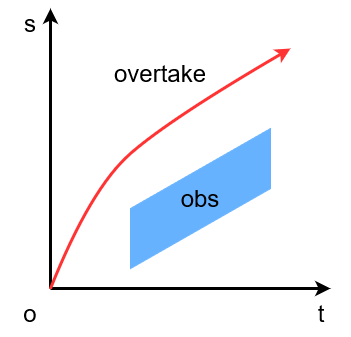}}
	\subfloat[\label{fig:st-b}]{
		\includegraphics[scale=0.3]{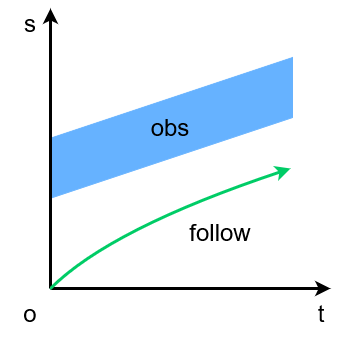}}
    \caption{ST graphs. (a) An obstacle prepares to cut in the current lane, so ADV chooses to overtake; (b) A preceding vehicle moves slowly, so the ADV slow down.}
    \label{fig:st-graph}
\end{figure}

Regarding following and overtaking maneuvers, these actions often arise when a preceding vehicle is moving at a slower pace or a new vehicle unexpectedly merges into a lane. The execution of these complex maneuvers is facilitated through the use of an s-t graph, which depicts the occupation of the lane by vehicles over time as illustrated in \cref{fig:st-graph}. By analyzing the s-t graph, end states can be sampled within unoccupied areas, marked in red (for overtaking) and blue (for following), thus facilitating the completion of these maneuvers through the use of a quintic polynomial.

Finally, these polynomials can be transformed back to Cartesian coordinate system to obtain the corresponding trajectories, which are consequently evaluated for compliance with kinematic constraints. If any trajectory does not meet the requirements, it will be removed from the sampling set.

\subsection{Deduction of GAN-based Generator}
\label{deduction}
Given the optimization target for evaluator:
\begin{equation}
    L_{evaluator} = \max_{\hat{s}}[E(\tilde{s}) - E(\hat{s}) + Diff(\tilde{s}, \hat{s})]
\end{equation}
The target for generator become:
\begin{equation}\label{e:gen}
    L_{generator} = E(\hat{s}) - E(\tilde{s}) + Diff(\tilde{s}, \hat{s})
\end{equation}
$E(\tilde{s})$ is determinate in the generator's step, thus omitted, and we expand $ Diff(\tilde{s}, \hat{s})$ as defined in \cref{eq:3} , \cref{e:gen} become:
\begin{equation}\label{e:e1}
    L_{generator} = E(\hat{s}) + L2(\tilde{s}, \hat{s}) + O(\hat{s}) + G_{rid}(\hat{s})
\end{equation}
Where:
\begin{equation}\label{e:expand}
    E(\hat{s}) = \sum C(\hat{s}) + \alpha O_{cc}(\hat{s}) + \beta G_{rid}(\hat{s})
\end{equation}
We then merge $O(\hat{s})$ and $G_{rid}(\hat{s})$ into \cref{e:expand}, tuned by hyper-parameter $\alpha$ and $\beta$, thus \cref{e:e1} become:
\begin{equation}
    L_{generator} = E(\hat{s}) + L2(\tilde{s}, \hat{s})
\end{equation}
Here the $L2$ is replaced by multi-path imitation defined in \cref{eq:6} to ensure diversity, the $\hat{s}$ here is given by the generator in \cref{eq:gang}, thereby:
\begin{equation}
    L_{generator} = E(G(m)) + L_{imit\_plan}(\tilde{s}, G(m))
\end{equation}

\subsection{Limitation of Open-loop Evaluation}
\label{app:limitation}
We illustrate the limitation of open-loop evaluation.
Firstly, a closed-loop planner will adjust to the target in the loop to keep safe and comfortable since real-world planning needs handle the deviations brought by the external disturbances and controller's modeling errors, which is invisible and omitted, thus cannot be tested by the open-loop planner. Secondly, the validity of the trajectory cannot be verified from the perspective of the downstream controller. The open-loop tests will come to an end once the trajectories are generated and compared with the recorded ones. However, unexpected and unreasonable jerks and curvatures, which can only be tested when the controller is in the loop, are usually ignored; therefore, the results become less convincing.
Existing imitation-based approaches formulate the modeling in planning task as prediction problem, which takes ego motion and scene feature to predict future movement. Predominantly, these models are assessed using imitation metrics and subjected to open-loop testing. Nevertheless, a fundamental discrepancy remains, as the aim of planning transcends mere imitation. 
It is imperative to meticulously craft the learning objectives to avert the entrapment of the model within a local optimum that, while demonstrating proficient motion prediction capacities, fails to adequately fulfill the planning objectives.

\subsection{Hardware}
Our autonomous driving algorithms have undergone rigorous testing on the recently launched Jiyue 01 vehicle, whose detailed parameters can be found in the following link: \url{https://www.jiyue-auto.com/robocar-01}. The Jiyue 01 is equipped with a comprehensive sensor suite, which includes 11 cameras strategically positioned to provide front, side, rear, and surround-view coverage. Additionally, the vehicle features 5 millimeter-wave radars, enabling accurate perception in challenging weather conditions, and 12 ultrasonic radars for close-range detection.

Moreover, the Jiyue 01 is fitted with 2 NVIDIA Orin-X chips, representing a powerful and advanced autonomous driving hardware and software platform. Specifically engineered for autonomous driving, these chips offer a processing capability of 508 TOPS, meeting the high demands of algorithm design and ensuring optimal performance. More NVIDIA-Orin parameters can be found in the following link: \url{https://developer.nvidia.com/drive/agx}.

\subsection{Task Perf}
\label{perf}

\begin{figure}[t]
    \centering
    \includegraphics[width=1\linewidth]{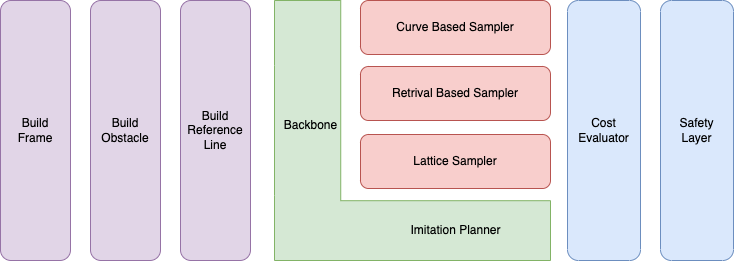}
    \caption{Tasks on board}
    \label{fig:on_board}
\end{figure}

\begin{table}[ht]
  \centering
  \begin{tabular}{c|ccc}
    \toprule
    Task &  \multicolumn{3}{c}{Time Cost(ms)} \\
    {} & Ave & 90th & Max \\
    \midrule
    Build Frame          &0.77  &1.13  &5.47\\
    Build Obstacles      &0.37  &0.67  &5.35\\
    Build Reference Line &0.52  &0.82  &10.61\\
    BackBone / Imitation planner &40.90  &51.05  &81.43  \\
    Curve Based Sampler  &5.12  &10.75 & 15.64 \\
    Retrival Based Sampler  &8.31  &12.16 &19.22  \\
    Lattice Sampler      &9.83  &15.15  &40.51  \\
    Cost Evaluator       &0.58  &1.39  &9.95\\
    Safety Layer         &0.32  &0.53  &8.12\\
    \midrule
    Over All             &50.24  &72.26 &108.72  \\
    \bottomrule
  \end{tabular}
  \caption{Task perf on board.}
  \label{tab:t4}
\end{table}
The perception module is implemented on the master chip, whereas the GAD resides on the subsequent chip. 
We enumerate the time expenditures associated with each task in \cref{tab:t4}. These tasks are orchestrated by a flow-based scheduler, facilitating concurrent execution as depicted in \cref{fig:on_board}. The deployment of the model via TensorRT is conducted without the utilization of acceleration techniques such as quantization or pruning.

\begin{figure*}[ht]
    \centering
    \includegraphics[width=1\linewidth]{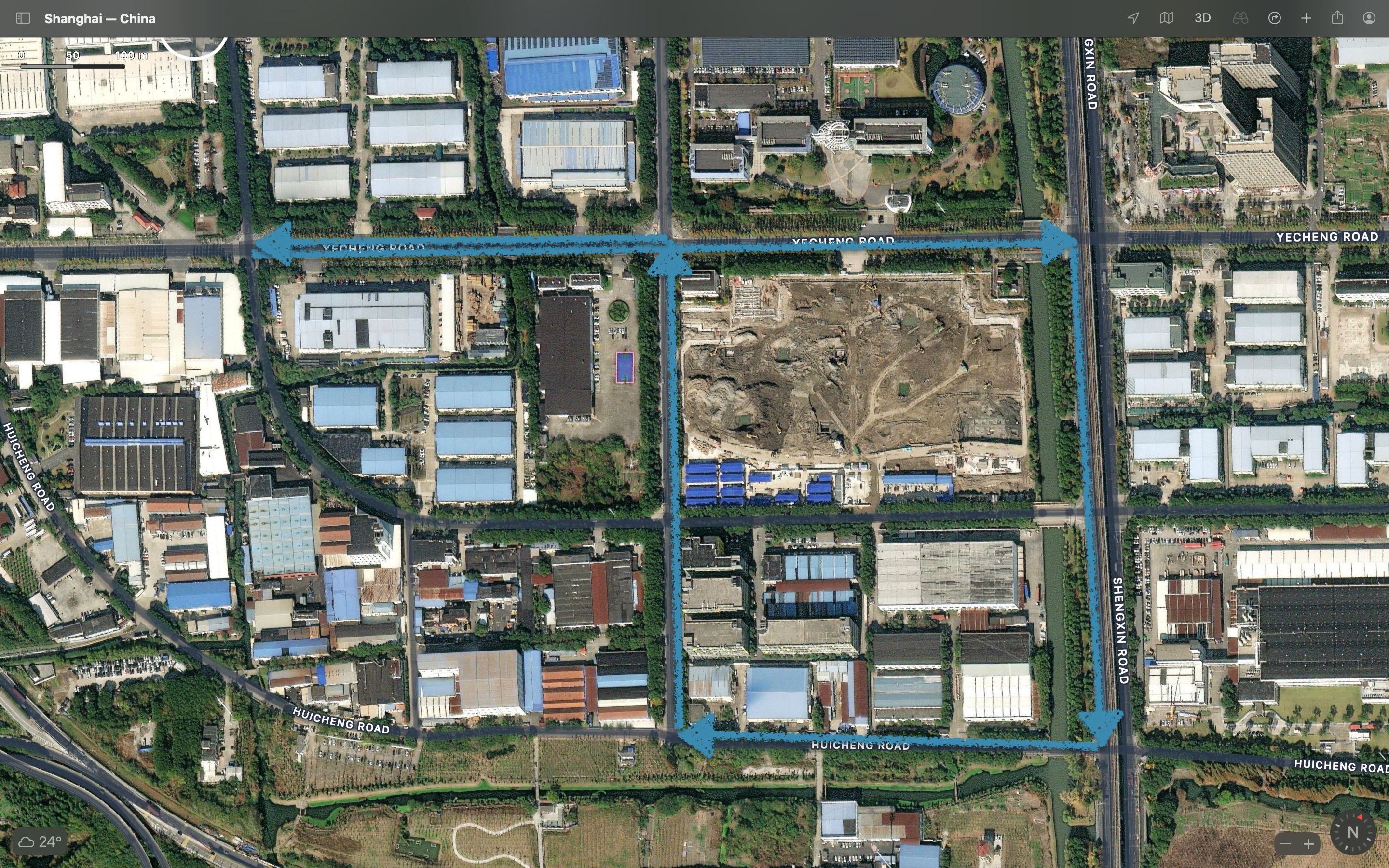}
    \caption{Test Drive Route}
    \label{fig:test_route}
\end{figure*}

\subsection{Road Test}
\label{roadtest}
Our road tests are conducted in the Jiading District of Shanghai, China, known as "Shanghai Automobile City". The detailed testing route is illustrated in Figure \ref{fig:test_route}, encompassing straight stretches, lane changes, intersections, and turning scenes involving various traffic participants.

The route begins at No.1688 Yecheng Road, a road with two lanes in each direction that encounters frequent cars and electric bicycles. It passes through two intersections controlled by traffic lights before entering a three-lane structure and making a careful lane change to the right-most lane, where sometimes occupied by static vans. It then queues before the stop line and then turns onto Shengxin Road, a four-lane-per-direction road heavily congested commuter route with significant traffic flows in all directions.

While weaving through vehicles on Shengxin Road and go straight across the first intersection, the route carefully enters the right-most lane, and then turns right, encountering electric bicycles and vehicles from different directions at the intersection. After turning onto South Huicheng Road, a one lane per direction road often occupied by passengers and tricycles, the traffic flows are slowed down, requiring careful navigations. 

The route then follows a right-turning at the next intersection and enters Huocheng Road, a narrow one-lane-per-direction road where trucks from the opposite direction sometimes encroaching on the ego lane, necessitating slight steering to avoid collisions. Proceeding through the next intersection, the route goes straight and executes an unprotected left turn to cross a bustling transportation route frequently used by trucks. Finally, the route circles back to Yecheng Road and returns to destination No.1688. The total length of the test route is about 3.0 kilometers long.

The video for road test is available at \url{https://youtu.be/DYPB5lCRvn0}

\subsection{Dataset}
\label{dataset}

We collected 15 hours of record files, containing messages such as perception outputs, ADV trajectories, traffic lights and route information. Following the collection, a thorough data cleaning process was implemented to analyze and identify the recorded data. Errors such as unconverged ADV localizations and perception frames of irregular frequencies were identified and excluded in order to enhance the quality and reliability of the dataset. The records were then categorized into different scenarios including cruising, lane-change, and turning scenarios.
Subsequently, less-common but crucial scenarios, such as unprotected left turning scenarios, were amplified by data augmentation, in order to achieve a well-balanced dataset. While data augmentation resulted in higher loss during training since difficult scenarios were added to the dataset, it enhanced the model's performance and generalization capability.
Finally, all the records were post-processed to obtain 200k clips in total. By default, we take the 1.5 seconds of past contexts and plan for the future 3.0 seconds of contexts, corresponding to 15 frames in the past and 30 frames in the future. We partition the dataset into 160k for training and 40k for testing.

\subsection{Training details}
\label{train_detail}
The Adam optimizer is used with an initial learning rate of $5e^{-4}$ and a batch size of 32. The scheduler of reduce on loss plateau decay is also implemented, with a reducing factor of 0.3 and a patience factor of 2, which means that if there is no improvement after 2 epochs, the learning rate will be reduced by 0.3 times  the original value. The lower bound of the learning rate is set to $10^{-8}$ and the model parameters are randomly initialized. 
Experiments are conducted with 4 NVIDIA A800 GPUs with 60 epochs to complete.

In the first stage of our proposed two-stage training model, we regress a grid map and cost volume by a CNN encoder-decoder backbone, which represents the occupation that vehicles may distribute in the future.
Specifically, CNN-encoder embeds the scene image into feature map with 12 convolution layers, and then 4 deconvolution layers are used to decode feature map to the output map. \par
In the second stage of obstacle trajectory prediction, we reuse the CNN encoding layer from the first stage and initialize the encoder parameters with the last training step. As for obstacle trajectory prediction, we use an LSTM layer to replace the CNN-decoder. 
In order to concentrate more on obstacles, we crop the encoding feature map centered agents, as the local scene feature of obstacles. The cropping direction aligns with the orientation of the obstacles. The local scene feature map will be embedding as a local scene vector through a Multi-Layer Perceptron layer and a max-pooling layer. Then, the local scene vector will be sent to the LSTM decoder, concatenated with agent's history vectorized embedding vector.

The training methodology for the GAN-based planner follow similar two-stage procedural, undergoing fine-tuning on the pre-trained imitation planner to stabilize and accelerate the training process.

\subsection{Qualitative Results}
\label{app:quali_resu}
\begin{figure*}[h]
    \centering
    \subfloat[\label{fig:qr-lkd1}]{\includegraphics[scale=0.3]{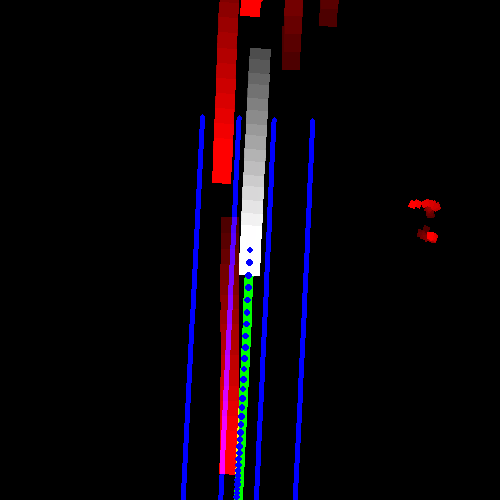}}\hspace{0.02\textwidth}
    \subfloat[\label{fig:qr-lkd2}]{\includegraphics[scale=0.3]{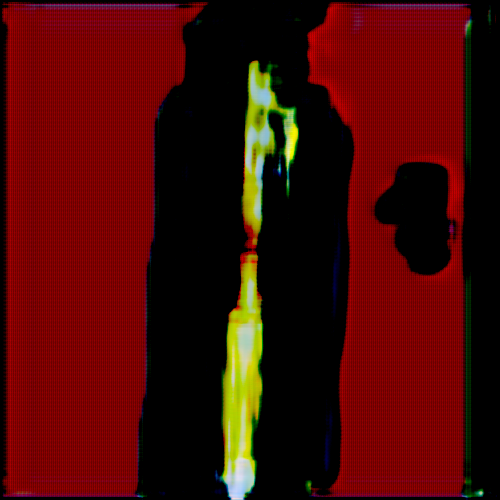}}\hspace{0.02\textwidth}
    \subfloat[\label{fig:qr-lkd3}]{\includegraphics[scale=0.3]{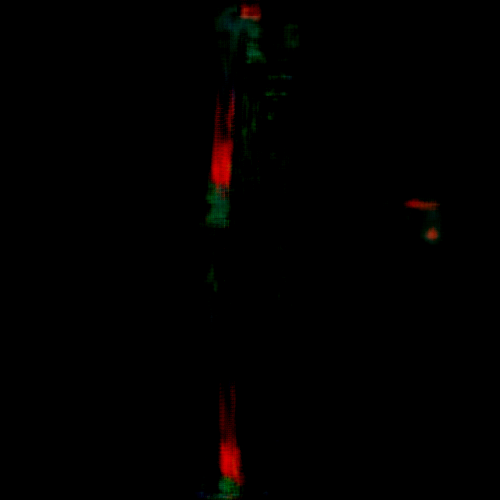}}
    \caption{Qualitative results: as the vehicle in the right lane accelerates and merges into the ego lane, the ADV opts to decelerate in order to prevent potential rear-end collisions.}
    \label{fig:qual_result}
\end{figure*}

\begin{figure*}[h]
    \centering
    \subfloat[\label{fig:qr-lc1}]{\includegraphics[scale=0.3]{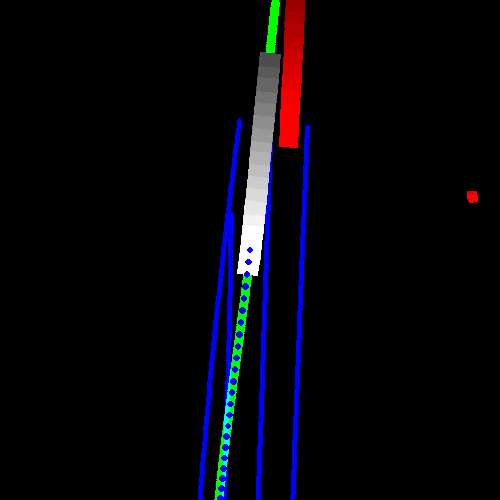}}\hspace{0.02\textwidth}
    \subfloat[\label{fig:qr-lc2}]{\includegraphics[scale=0.3]{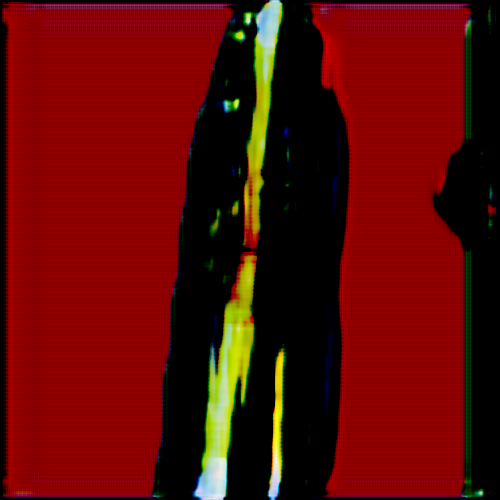}}\hspace{0.02\textwidth}
    \subfloat[\label{fig:qr-lc3}]{\includegraphics[scale=0.3]{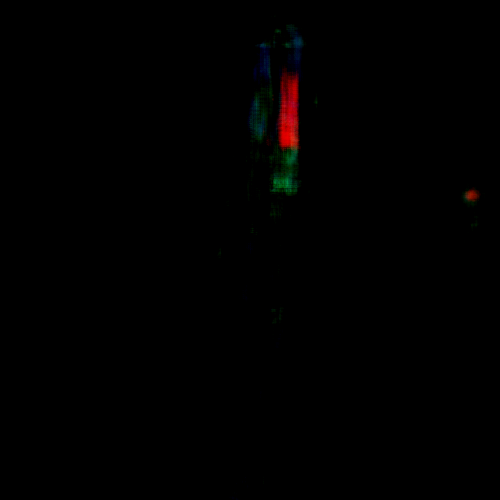}}
    \caption{Qualitative results: the ADV is transitioning to the rightmost lane, while also considering the option of using the middle lane if the rightmost lane is occupied.}
    \label{fig:changelane}
\end{figure*}

\begin{figure*}[h]
    \centering
    \subfloat[\label{fig:qr-tj1}]{\includegraphics[scale=0.3]{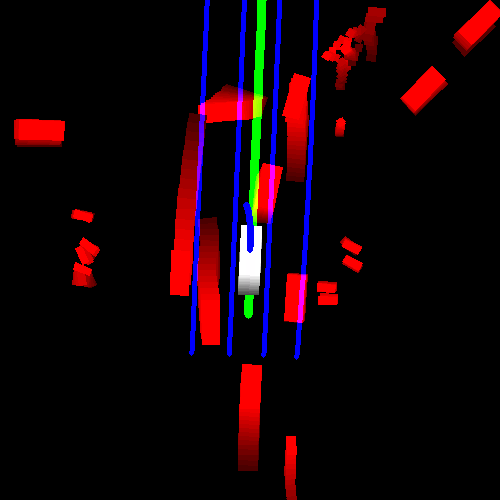}}\hspace{0.02\textwidth}
    \subfloat[\label{fig:qr-tj2}]{\includegraphics[scale=0.3]{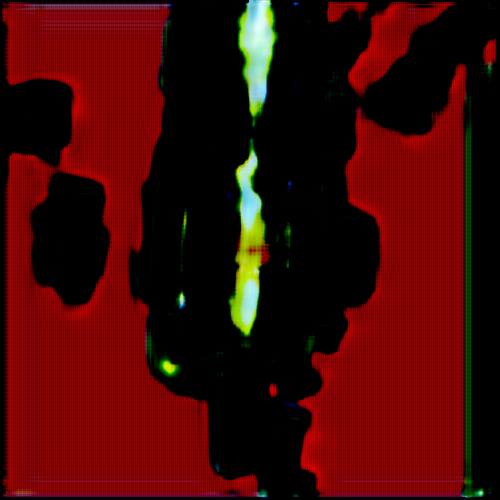}}\hspace{0.02\textwidth}
    \subfloat[\label{fig:qr-tj3}]{\includegraphics[scale=0.3]{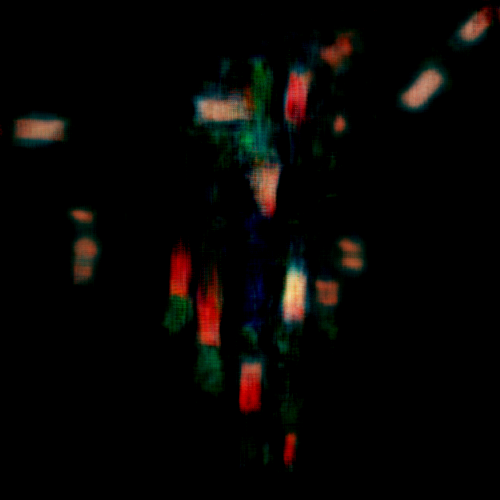}}
    \caption{Qualitative results: the ADV  is maneuvering out of the traffic congestion by overtaking the preceding vehicle at a low speed in order to mitigate potential hazards.}
    \label{fig:jam}
\end{figure*}

\begin{figure*}[h]
    \centering
    \subfloat[\label{fig:qr-t311}]{\includegraphics[scale=0.3]{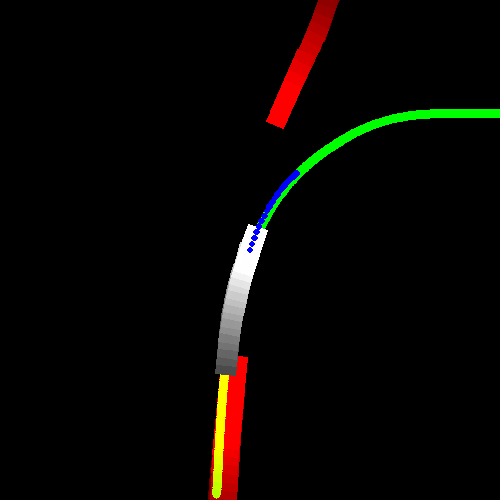}}\hspace{0.02\textwidth}
    \subfloat[\label{fig:qr-t312}]{\includegraphics[scale=0.3]{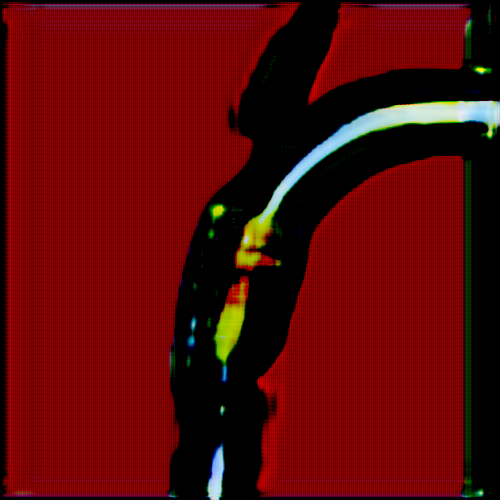}}\hspace{0.02\textwidth}
    \subfloat[\label{fig:qr-t313}]{\includegraphics[scale=0.3]{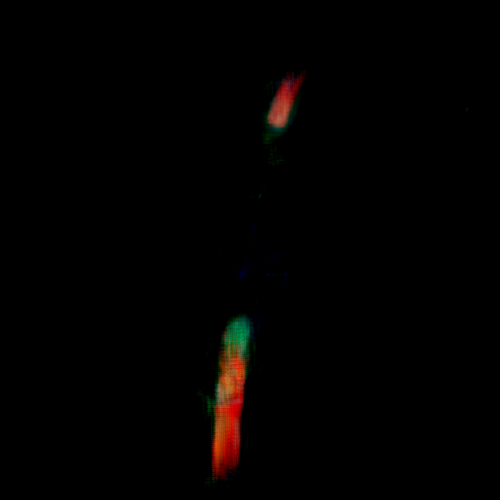}}
    \caption{Qualitative results: the ADV is performing a right turn at a decreased speed to prioritize safety.}
    \label{fig:turningright}
\end{figure*}

\begin{figure*}[h]
    \centering
    \subfloat[\label{fig:qr-t411}]{\includegraphics[scale=0.3]{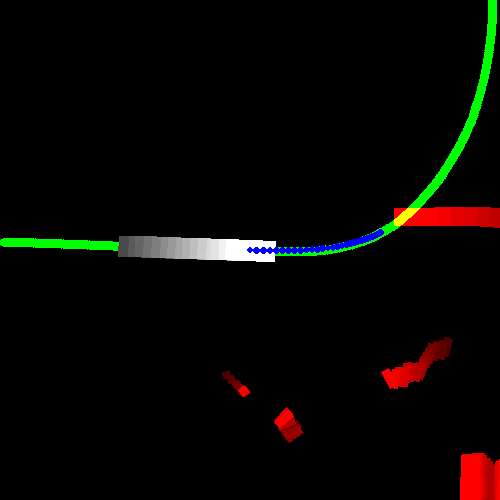}}\hspace{0.02\textwidth}
    \subfloat[\label{fig:qr-t412}]{\includegraphics[scale=0.3]{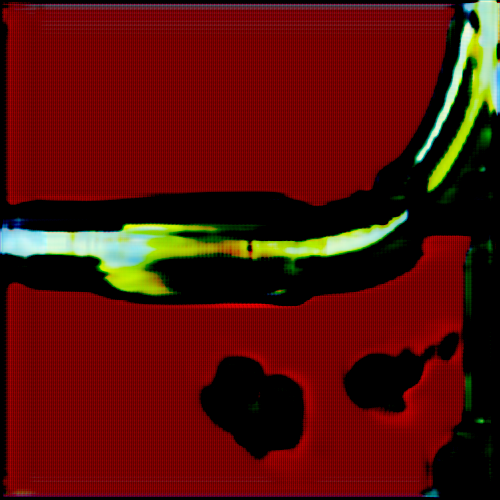}}\hspace{0.02\textwidth}
    \subfloat[\label{fig:qr-t413}]{\includegraphics[scale=0.3]{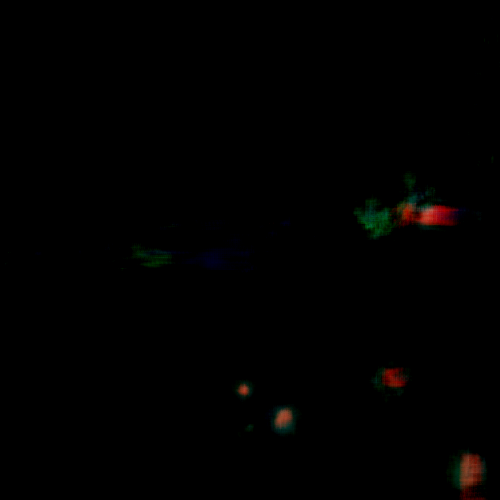}}
    \caption{Qualitative results: the ADV is making a left turn at an unprotected intersection. As a vehicle from the oncoming direction approaches, the ADV reduces its speed.}
    \label{fig:unprotectedturning}
\end{figure*}

We show typical qualitative results in \cref{fig:qual_result} - \cref{fig:unprotectedturning}  to illustrate that our work ensures the ADV can navigate through diverse traffic scenarios to reach the destination safely. Note that (a) shows the traffic scenario, and the blue points are the final trajectory selected by our model. (b) is the visualization of the cost volume which explains the areas with higher probabilities to travel. (c) is the visualization of the grid map to illustrate the future occupancies of obstacles.
It should be noted that, figure (b) shows the superposition of our predictions for the occupancy at the next 1s, 2s, and 3s. The red area represents the highest probability of occupancy for the next 1s for the ego vehicle. The green and blue areas represent the highest probability of occupancy for the next 2s and 3s, respectively.  Figure (c) illustrates the superposition of occupancy at the current moment, in 1s, and in 2s. The red region denotes the current position of the obstacle, while the green indicates the areas that the obstacle may occupy at 1s, and the blue represents the areas the obstacle may occupy at 2s.

\end{document}